\pdfoutput=1

\documentclass[11pt]{article}

\usepackage[preprint]{acl}

\usepackage{times}
\usepackage{latexsym}

\usepackage[T1]{fontenc}

\usepackage[utf8]{inputenc}

\usepackage{microtype}

\usepackage{inconsolata}

\usepackage{graphicx}

\usepackage{booktabs}
\usepackage{amsmath}
\usepackage{hyperref}
\usepackage{url}
\usepackage{times}
\usepackage{latexsym}
\usepackage[T1]{fontenc}
\usepackage{float}
\usepackage{fontawesome}

\usepackage{placeins}
\usepackage[utf8]{inputenc}
\usepackage{microtype}
\usepackage{inconsolata}
\usepackage{subfig}
\usepackage{amsmath}
\usepackage{subcaption}
\usepackage{amssymb}
\usepackage{booktabs}
\usepackage{xspace}
\usepackage{xcolor}
\usepackage{multicol}
\usepackage{multirow}
\usepackage{array}    

\usepackage{enumitem}
\usepackage{graphicx}
\RequirePackage{algorithm}
\RequirePackage{algorithmic}

\title{Merging Feed-Forward Sublayers for Compressed Transformers}

\author{
    Neha Verma$^{1}$ \qquad Kenton Murray$^{1,2}$ \qquad Kevin Duh$^{1,2}$ 
    \\
    $^{1}$Center for Language and Speech Processing \\
    $^{2}$Human Language Technology Center of Excellence \\
    Johns Hopkins University \\
    {\tt \{nverma7, kenton\}@jhu.edu, kevinduh@cs.jhu.edu}\\
}

\begin{document}
\maketitle
\begin{abstract}
With the ubiquity of large deep learning models and their growing number of use cases, the need for high-quality compression techniques is growing in order to deploy these models widely across diverse hardware and memory settings. 
In this work, we present a novel approach to model compression by \textit{merging} parameter groups within a model, rather than pruning away less important parameters.
Specifically, we select, align, and merge separate feed-forward sublayers in Transformer models, and test our method on language modeling, image classification, and machine translation. 
With our method, we demonstrate performance comparable to the original models while combining more than a third of model feed-forward sublayers, and demonstrate improved performance over a strong layer-pruning baseline. 
For instance, we can remove over 21\% of total parameters from a vision transformer, while maintaining 99\% of its original performance. 
Additionally, we observe that some groups of feed-forward sublayers exhibit high activation similarity, which may help explain their surprising mergeability. 
\end{abstract}

\section{Introduction}

Recent advances in deep learning have been marked by large, pre-trained models in order to achieve state-of-the-art performance. 
As these models deploy across a wider range of settings, compression techniques that balance efficiency and performance are increasingly important.
These techniques help facilitate model use across a variety of inference settings and hardware availability.

Much of the prior work in model compression has built upon on pruning, quantization, and distillation techniques \citep{lecun1989optimal, fiesler1990weight, hinton2015distilling}. 
Prior work on pruning has introduced many techniques identifying regions of model parameters that can be removed without drastically changing performance. 
These techniques target individual weights, neurons, or general regions of a model---like attention heads,  parameter chunks, or even entire layers. \citep{voita-etal-2019-analyzing, lagunas2021block, sajjad2023effect}.
However, while ``unimportant'' features are targeted for pruning techniques, we can also target ``redundant'' features for compression. 
There has been far less focus on compression methods that target redundancy within a model.

In targeting redundant features for compression, we can \textit{merge} sets of similar parameters rather than prune them. 
Relatedly, the field of model merging has explored merging parameters from multiple models to combine their functionalities into one model \citep{goddard2024arcee, yang2024modelmergingllmsmllms}. 
We instead extend parameter merging to \textit{sublayers} within one model, rather than just separate models. 

To this end, we propose a novel compression method that aligns, merges, and ties separate feed-forward (FF) sublayers within Transformers to achieve a reduced parameter model with reduced memory use \citep{Vaswani2017AttentionIA}. We target FF sublayers in particular due to their large parameter count and easy mergeability.  
Through our testing, we find that these groups of FF sublayers are notably compressible via merging, giving rise to a simple and surprisingly effective framework applicable to a variety of existing pre-trained models. 

 \begin{figure*}[t]%
    \centering
    \includegraphics[width=\textwidth]{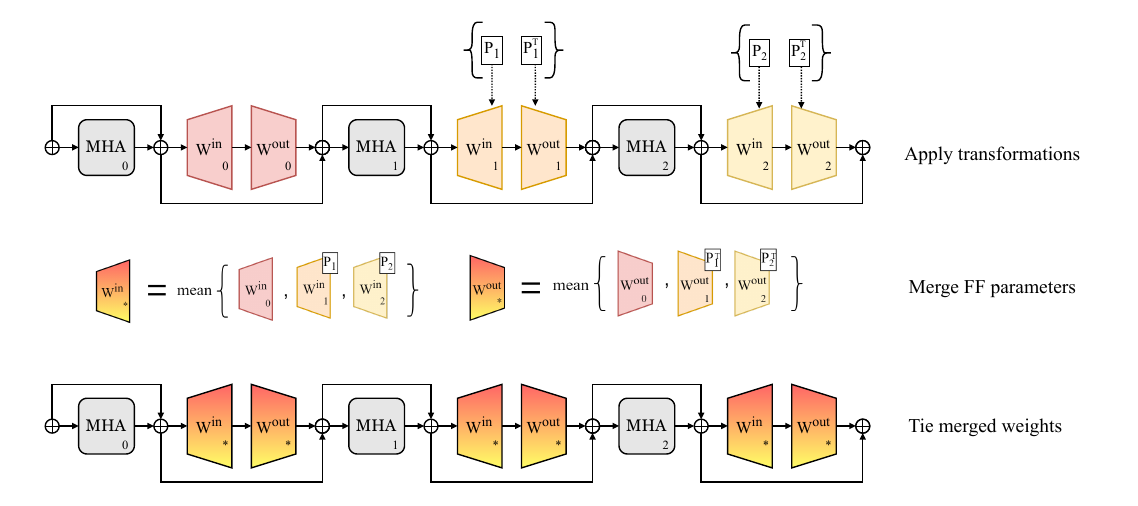}
    \caption{Overview of the feed-forward alignment and merging algorithm used to compress models in an example three layers of a Transformer. Multi-headed attention is abbreviated to MHA, feed-forward sublayers are depicted with $W^\text{in}$ and $W^\text{out}$ weights, and Add\&Norm operations are depicted with $\bigoplus$, connected by arrows indicating residual connections.\protect\footnotemark~ Permutation transformation matrices are shown as $P_i$. Our method includes a permutation finding step, applying the transformations, merging transformed parameters, and finally tying the merged parameters. By merging and tying $k$ feed-forwards, we can reduce the model size by $k-1$ feed-forward sublayers.}
    \label{fig:method_overview}
\end{figure*}

We highlight the contributions of our work:
\begin{itemize}[topsep=0pt,itemsep=0pt]
    \item We propose a novel model compression method inspired by recent work in model merging. This approach is orthogonal to compression methods like quantization. 
    \item Across three different Transformer-based models, namely GPT-2, ViT, and a machine translation model, we show that merging over one-third of FF sublayers and fine-tuning the resulting model can achieve performance comparable to the original models. We also combine our method with QLoRA to help facilitate even smaller fine-tuning settings \cite{dettmers2023qlora}.
   
    \item To explore the surprising effectiveness of merging, we compare different FF outputs from the same model, and find regions with highly similar activations. These same patterns do not occur in attention outputs. 
    \item We release an easily extensible toolkit for our method: \fontsize{10.5}{9}\selectfont\href{https://github.com/nverma1/merging-ffs-compression}{\faGithub ~nverma1/merging-ffs-compression}
\end{itemize}
\footnotetext{This diagram shows a Post-LN Transformer, but our method easily applies to Pre-LN Transformers as well. }

\vspace{-0.5em}
\section{Related Work}

In this section, we review prior work related to weight tying and redundancies in Transformers.

\paragraph{Weight tying for smaller models} Prior work on weight tying has focused on training models from scratch with specific tying schemes. 
Tying input and output embedding layers helps cap total parameter count, but more importantly introduces important gradient sharing for better generalization in language generation tasks \citep{press2017using, inan2017tying}.
For non-embedding layers in Transformers, prior work has explored numerous weight tying patterns for pre-training new, efficient models \citep{dehghaniuniversal, Lan2020ALBERT, reid-etal-2021-subformer-exploring, takase-kiyono-2023-lessons}. 
\citet{liumobilellm} use heavy weight tying between Transformer layers at initialization to achieve state-of-the-art sub-billion parameter language models.
\citet{pires-etal-2023-one} tie widened FF sublayers at initialization and train machine translation (MT) models that outperform standard Transformer MT models. In our work, we instead start from a pre-trained model, and then use weight sharing as a \textit{post-training} tool to reduce overall parameter count. 

\paragraph{Redundancies in Transformers} Prior work has demonstrated signs of redundancy in Transformer computations, and suggested techniques to reduce this phenomenon.  
\citet{dalvi-etal-2020-analyzing} use centered kernel alignment (CKA) to show high layer similarity in BERT and XLNet, and use correlation clustering to find and remove redundant neurons.
\citet{men2024shortgpt, gromov2024unreasonable} propose removing Transformer layers in deep, decoder-only language models to achieve faster inference at a small performance drop. 
\citet{limerge} propose a compression method for sparsely-activated mixture-of-expert (SMoE) models that draws from model merging to compress experts in  SMoE models. Our method extends a similar approach to a much wider set of models.

\section{Merging Feed-Forward Sublayers}

In this section, we discuss FF sublayers as a merging target, explain permutation-based neuron alignment, and describe our compression method.

\subsection{Targeting feed-forward sublayers}

We focus our interest on Transformer FF sublayers for several reasons. 
Firstly, these sublayers constitute around two-thirds of non-embedding parameters in Transformer encoder or decoder models.
Compressing these parameters results in substantial overall savings in a model.
Secondly, the parameterization of FF sublayers, including variations, is straightforward, making it a good candidate for merging-based compression approaches.

Beyond these practical considerations, prior work establishes several properties of Transformer FF sublayers that make them good candidates for compression via merging. 
\citet{lilazy} show that they can be very sparsely activated, where non-zero FF activations can be as low as 3-5\%. 
Other work has demonstrated evidence that adjacent LayerNorm and FF blocks, in both Post- and Pre-LN architectures, results in weakening of the contextualization effects of FF sublayers \citep{kobayashianalyzing}. 
The authors allude to redundancy in Transformer FF processing due to this interaction.
Finally, \citet{pires-etal-2023-one} train performant Transformer-based translation models with only one widened and tied encoder FF block, demonstrating useful sharing, but from scratch.

\subsection{Background on permutation-based neuron alignment}
\label{sec:perm_background}
We propose a merging technique that combines several similar sublayers into a single parameter set.
Our merging technique is inspired by prior work in permutation symmetries of neurons \citep{pmlr-v44-li15convergent}.
This technique has been used in studying models' convergent learning, as well as merging two or more separate models \citep{tatro2020optimizing, entezari2021role, ainsworthgit}. 

Permutation-based alignment techniques seek to find an optimal reordering of neurons in one layer that more closely matches ordering of neurons from another layer, without changing the output. 
Given two layers to align, we compute only forward passes through both using exemplar data to collect activations.
These layers are generally corresponding from different models.
This results in two activation sets $X_{\alpha}, X_{\beta} \in \mathbb{R}^{n \times d}$, where $n$ is the number of example data points, and $d$ is the model dimension.

To determine corresponding neurons from the activations, we compute cross-correlation $C$, in line with prior work \citep{pmlr-v44-li15convergent}. $\mu$ represents mean vectors, and $\sigma$ standard deviation vectors. 
\begin{equation}
    C  = \frac{\mathbb{E}\left[\left(X_{\alpha} -  \mu(X_{\alpha})\right)^T\left(X_{\beta} - \mu(X_{\beta})\right)\right]}{\sigma(X_{\alpha})\sigma(X_{\beta})} 
\end{equation}
The resulting matrix $C \in \mathbb{R}^{d \times d }$ reflects how each neuron $j$ in $X_{\alpha}$ correlates with each neuron $k$ in $X_{\beta}$.
To find the neuron alignment that maximizes total correlation, we solve the following optimization problem, where $\Pi_d$ is the space of all permutations of length $d$ \citep{pmlr-v44-li15convergent, tatro2020optimizing}:
\begin{equation}
    \pi^* = \max_{\pi \in \Pi_d} \sum_{j=1}^d C(j, \pi(j))
\end{equation}
This problem is a case of the Linear Assignment Problem, and we solve for $\pi^*$ using the Jonker-Volgenant algorithm implementation provided by $\texttt{scipy}$ \citep{7738348}. 

\subsection{Combining feed-forward sublayers}
\label{sec:main_method}
 
Now, with the appropriate background, we describe our compression method.
For our method, we assume we have some predetermined number of FF sublayers $k$ to merge. This number can be inferred given a desired parameter reduction ratio, or set otherwise.

Given a window of $k$ adjacent FF sublayers, we compute a forward pass using a subset of data in order to compute features for each sublayer.
In other words, for Transformer FF sublayer $x^{\text{out}} = W^{\text{out}}\phi(W^{\text{in}}x^{\text{in}} + b^{\text{in}}) + b^{\text{out}}$, we obtain features just before the $\phi$ activation. 
We consider only the neurons just \textit{after} $W^{\text{in}}$ because prior work has shown that to reorder the input to $W^{\text{in}}$ and output of $W^{\text{out}}$ requires permuting many additional weights due to the residual connections in order to maintain functional equivalence \citep{verma2024mergingtexttransformermodels}. 
For each of the $k$ feed-forward sublayers, we collect features $X_i \in \mathbb{R}^{n \times d}, ~~ i \in [0,k-1]$, where $d$ is the feed-forward dimension.\footnote{The layer indices reflect local index within the set of $k$ versus global layer index.}

We designate the first FF sublayer of the set to be an ``anchor'', and compute the permutation-finding algorithm on each pair of features where one index is always the anchor. In other words, for each sublayer $i \in [1, k-1]$, we find $\pi_{i}$ between $X_0$ and $X_i$ using the assignment method from Section \ref{sec:perm_background}. 

After converting function $\pi_i$ to its corresponding permutation matrix $P_i$, we transform the $k-1$ non-anchor FF sublayers.
 We then average these $k$ FF sublayers, and replace each of them with their average, as in Equations \ref{eq:in}--\ref{eq:out}.\footnote{For an extension of this method to SwiGLU \cite{shazeer2020glu}, see Appendix \ref{app:swiglu}.} Finally, we tie these weights so that in memory they appear as just one sublayer, effectively removing the parameters from $k-1$ FF sublayers. 
\begin{align}
    \label{eq:in}
    W^{\text{in}*} &= \frac{1}{k} \left(W_0^{\text{in}} + \sum_{i=1}^{k-1} P_i W_i^{\text{in}} \right)  \\
    b^{\text{in}*} &= \frac{1}{k}\left(b_0^{\text{in}} + \sum_{i=1}^{k-1} P_i b_i^{\text{in}} \right) \\
    W^{\text{out}*} &= \frac{1}{k} \left(W_0^{\text{out}} + \sum_{i=1}^{k-1} W_i^{\text{out}}P_i^T\right) 
    \\ \label{eq:out}
    b^{\text{out}} &= \frac{1}{k} \left(\sum_{i=0}^{k-1} b_i^{\text{out}} \right)
\end{align}

\paragraph{Practical implications} The consequences of tying these weights and reducing model size are 1) the model occupies less space on disk or on GPU when fine-tuning (as well as its gradients), making it easier to use smaller hardware and 2) if some optimizations are made, inference speed and throughput can be improved. While naively our method does not target inference speed, we discuss ways this could be achieved. For example, if larger batch sizes are used given the model memory savings, this can result in increased throughput on the same hardware. Additionally, specific efficient GPU+CPU execution techniques like layer-to-layer that involve on- and off-loading parameters may also benefit from this sharing scheme by reducing the number and size of data transfers \cite{pudipeddi2020training, aminabadi2022deepspeed}.
\subsection{Selecting sublayers to merge}
\label{sub:select_layers}

In selecting the $k$ adjacent feed-forward sublayers to merge, we take a sliding window approach. 
For all starting layer indices from $0$ to $(N_{\text{layers}} -1) -k $, we apply the method outlined in Section $\ref{sec:main_method}$, and evaluate the resulting model on a validation set. 

Although we propose to test each potential window, in reality, the cost of computing permutations and parameter arithmetic is low, and it scales only linearly with the number of layers. 
The largest costs in each iteration is computing features and testing candidates. 
However, we can compute features only \textit{once} despite testing $N_{\text{layers}} -k $ models, because one forward pass through the exemplar data is sufficient for creating all necessary correlation matrices.  
The best candidate is the one with the highest post-merge evaluation score. 
We note that there may be other possible selection heuristics.

Finally, we follow our merging procedure with a short recovery fine-tuning  to quickly heal performance on the downstream task.\footnote{An algorithm summarizing our selection method can be found in Appendix \ref{app:algo_select}.}

\section{Experimental Setup}

To test its extensibility, we apply our compression method to a diverse set of Transformer-based models.
Specifically, we use GPT-2 \citep{radford2019language}, a vision transformer (ViT) \citep{dosovitskiy2020image}, and a Transformer-based MT model from OPUS-MT \citep{TiedemannThottingal:EAMT2020}. We select these models in order to cover a diversity of Transformer model types (decoder-only, encoder, encoder-decoder) and different modalities. We additionally include experiments on OLMo-7B using QLoRA \cite{groeneveld2024olmo}.

For each main setting, we list the model used, the example data for computing alignments, and finally the data used for recovery fine-tuning and evaluation. Additional hyperparameters are included in Appendix \ref{app:ft_details}, and dataset details in Appendix \ref{app:data_details}.

\subsection{Language modeling}

For our experiments, we use GPT-2 Large which was trained on extensive English text \cite{radford2019language}. 
For computing example activations, we use \textasciitilde10k tokens from the validation set of the Wikitext103 dataset \citep{merity2017pointer}. 
Finally, we use the train and test sets from the Wikitext103 for fine-tuning and evaluation, respectively. 

Because we use Wikitext103 for recovery fine-tuning, we also fine-tune the uncompressed GPT-2 baseline model for fair comparison. Because we have access to the training data for our machine translation and ViT models, we do not provide a fine-tuned baseline for those as the data we use already appears in their original training data.

We fine-tune our GPT-2 models for up to 100k steps with batches of 2048 tokens. We select the best model based on validation perplexity and report average test perplexity with a sliding window of 512 tokens. 

\subsection{Image classification with ViT}

We use a vision transformer (ViT) for our image classification experiments, with resolution of 224x224, and patch size of 16x16 \cite{dosovitskiy2020image}.
ViT is a 12-layer Transformer encoder model pre-trained on ImageNet-21k, and subsequently fine-tuned on ImageNet-1k. 
ImageNet-1k is a classification task where images belong to one of 1000 categories \citep{imagenet15russakovsky}.
For computing activations, we use \textasciitilde10k patches from the ImageNet-1k validation set. Evaluation results are computed on original validation labels. 

We fine-tune our ViT models on ImageNet-1k for up to 50k steps with a batch size of 128, and report accuracy.

\subsection{Machine translation}

For our experiments on machine translation, we use a 12-layer Chinese-English Transformer-based translation model from an OPUS-MT release \citep{TiedemannThottingal:EAMT2020}.
For computing activations, we use \textasciitilde10k tokens from the Tatoeba validation set\footnote{Tokens are counted on the source side.} \citep{tiedemann-2020-tatoeba}. 
For fine-tuning, we use the original training data released by the Tatoeba translation challenge, sourced from OPUS \citep{tiedemann-2012-parallel}.

We apply our method to both the encoder and decoder separately, constituting two anchors. However, we search windows in sync, meaning that the same window from the encoder and decoder are merged, but separately. 

We fine-tune our translation models for up to 100k steps with a batch size of 64 sentences. 
We use \texttt{sacrebleu} to compute BLEU scores for evaluation \citep{papineni-etal-2002-bleu, post-2018-call}, and \texttt{pymarian} to compute COMET scores\footnote{We use the \texttt{wmt-22-da} model.}  \citep{rei-etal-2022-comet, gowda2024pymarian}.

\subsection{Layer pruning baseline}

Recent work on the structured pruning of Transformers has seen many methods that remove full layers from a model and then optionally fine-tune the compressed model \citep{ma2023llm, men2024shortgpt, gromov2024unreasonable, yang2024laco}. We focus on a structured pruning baseline as many unstructured pruning methods do not realize memory savings unless they achieve 1) high sparsity ratios and 2) use specialized sparse libraries to store sparse weights. On the other hand, our method easily realizes compression due to weight tying.

We implement layer-pruning as a baseline; many layer-pruning methods rely on similarity measures to choose a set of adjacent layers to prune. However, we avoid any specific similarity techniques and instead choose the best subset after evaluation much like our own technique, via a sliding window and testing on validation data. After selecting the best pruned model, we then fine-tune the model with the same specifications as our method. In all, this encapsulates a strong, structured pruning baseline that generalizes many layer-pruning based techniques.

 \begin{figure*}[t]%
    \centering
    \subfloat[\centering ViT]{{\includegraphics[width=0.3\textwidth]{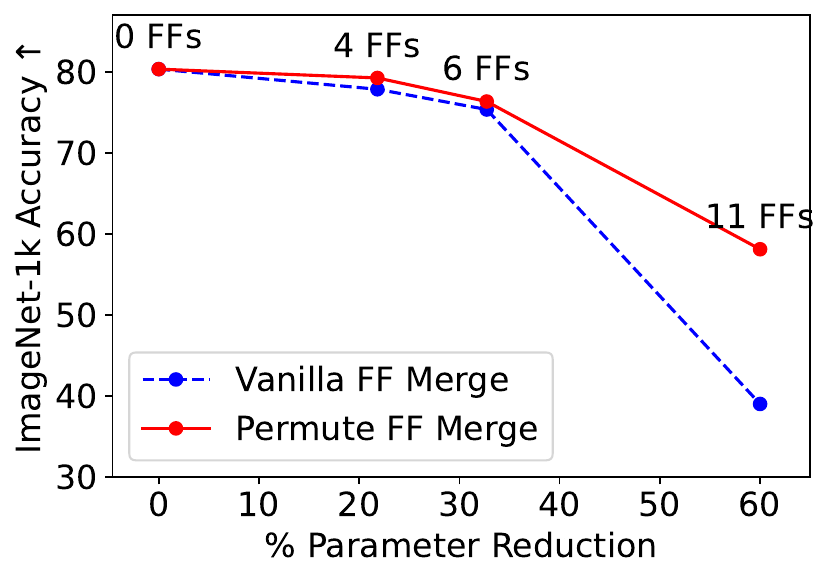} }}
    \subfloat[\centering GPT-2]{{\includegraphics[width=0.3\textwidth]{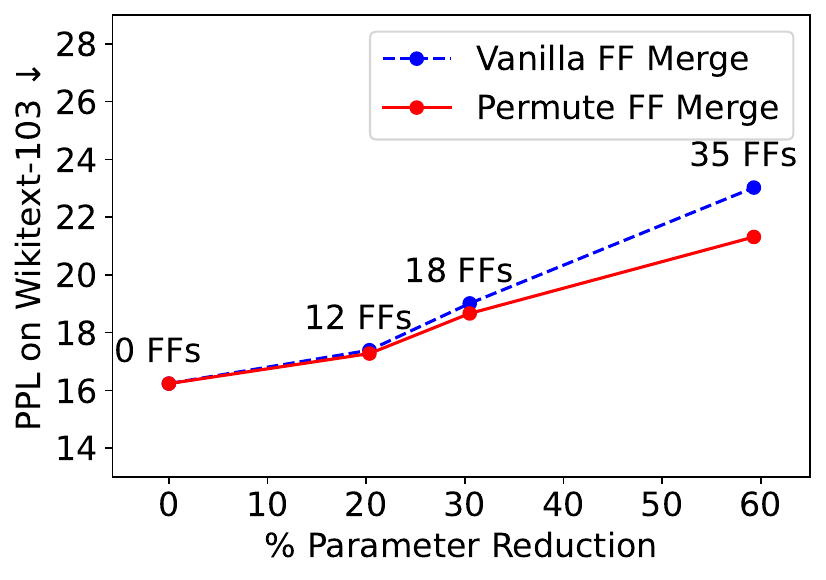} }}%
    \subfloat[\centering OPUS-MT]
    {{\includegraphics[width=0.3\textwidth]{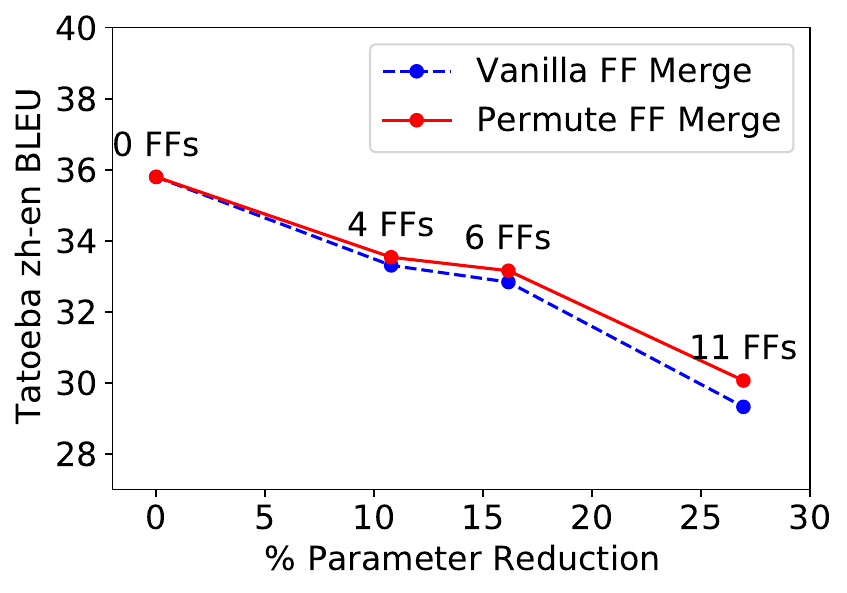} }} \\
    \caption{Results across all three tasks depicting compression versus performance results. We include results from our main method, labeled as Permute FF Merge, as well as our method without permutation alignment, depicted as Vanilla FF Merge. We note that our method retains almost complete performance at one-third of feed-forward sublayers removed, across all tasks, and continues to retain high performance at one-half of FF sublayers removed.}
    \label{fig:main_curves}
\end{figure*} 

\subsection{OLMo-7B QLoRA Extension}

We additionally apply our method alongsize 4-bit QLoRA to OLMo-7B for a downstream summarization task, namely SamSum \cite{groeneveld2024olmo, gliwa-etal-2019-samsum}. SamSum is an English dialogue summarization dataset, and OLMo-7B is a 32-layer English language model trained on the open Dolma dataset \cite{, soldaini-etal-2024-dolma}. This model uses SwiGLU FFs; we discuss extending our method to this variation in Appendix \ref{app:swiglu}. We compute features on \textasciitilde10k tokens from Dolma, and select the best pre-tune model using Wikitext-103. We report results on SamSum using ROUGE-1, ROUGE-2, and ROUGE-Lsum \cite{lin-2004-rouge}.

\section{Results}

\subsection{Merging feed-forward sublayers across compression ratios}

We evaluate our compression method on image classification using ViT, language modeling using GPT-2, and machine translation using an OPUS-MT zh-en model, and report our results in Figure \ref{fig:main_curves}. We report results at 1/3, 1/2 and $(n-1)/n$ feed-forward sublayers removed, in order to test our method at different overall compression ratios.\footnote{Compression ratios for OPUS-MT differ due to the enc-dec architecture.} 
We also report results from our compression method without the permutation step, labeled as ``Vanilla.''

From our results, we see that even up to 1/2 of FF sublayer parameters removed, which is over 30\% in parameter reduction for ViT and GPT-2,\footnote{We include embedding parameters in all \% parameter reduction and compression ratio calculations.} our method can retain high performance.
At 1/3 of FF sublayers removed, performance is almost identical to the original model, resulting in only a 1\% accuracy drop in ViT, 1 PPL increase in GPT-2, and 2 BLEU drop in the translation model. 
Full numerical results can be found in Appendix \ref{app:full_results}.

Our findings also hold across all three of our tasks tested, suggesting that our method generalizes to different types of models.
Additionally, we can notice that permutation-based compression is consistently better compared to no-permute vanilla baselines, demonstrating the effectiveness of aligning features before merging. This effectiveness is more pronounced at larger numbers of FF sublayers removed. In summary, our results show that 1) post-training weight tying is a simple and effective compression method and 2) permutation-based alignment of these shared weights can improve final compression performance.

 \begin{figure*}[h!]%
    \centering
    \subfloat[\centering ViT]{{\includegraphics[width=0.3\textwidth]{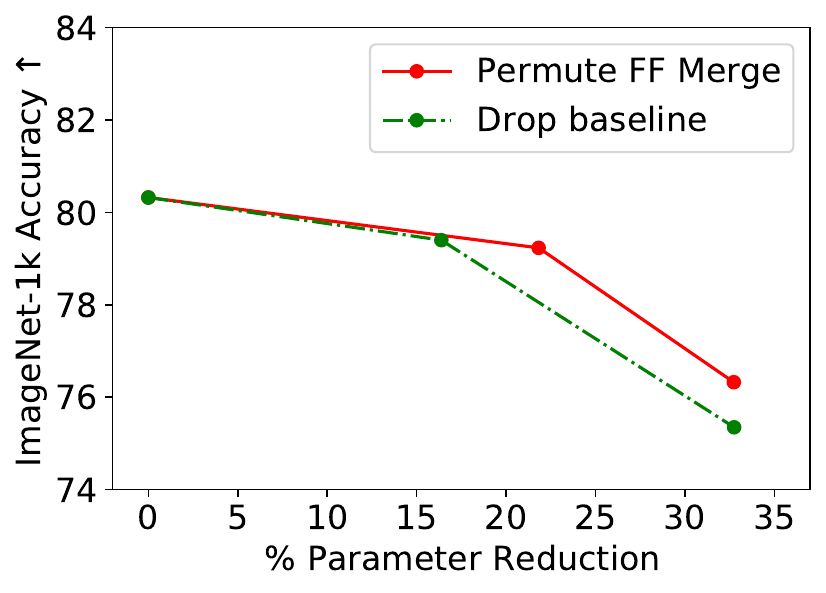} }}
    \subfloat[\centering GPT-2]{{\includegraphics[width=0.3\textwidth]{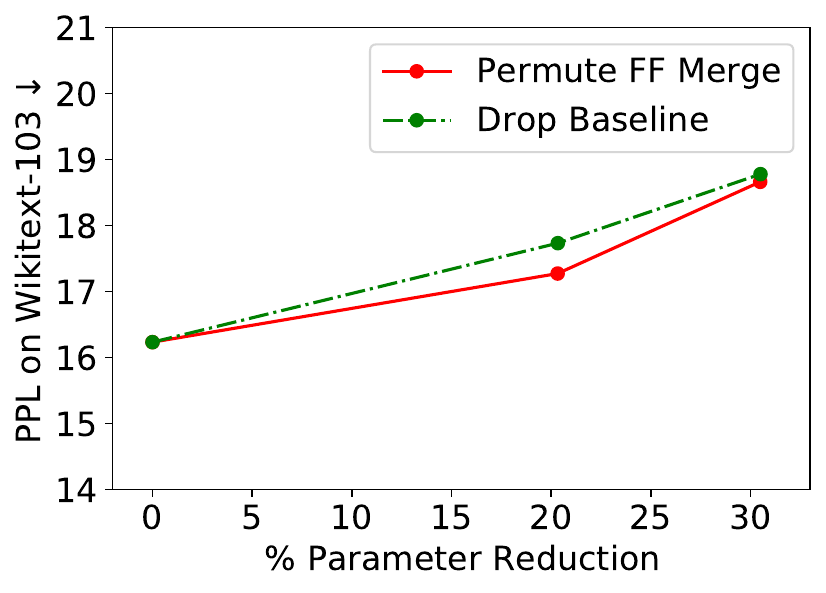} }}%
    \subfloat[\centering OPUS-MT]
    {{\includegraphics[width=0.3\textwidth]{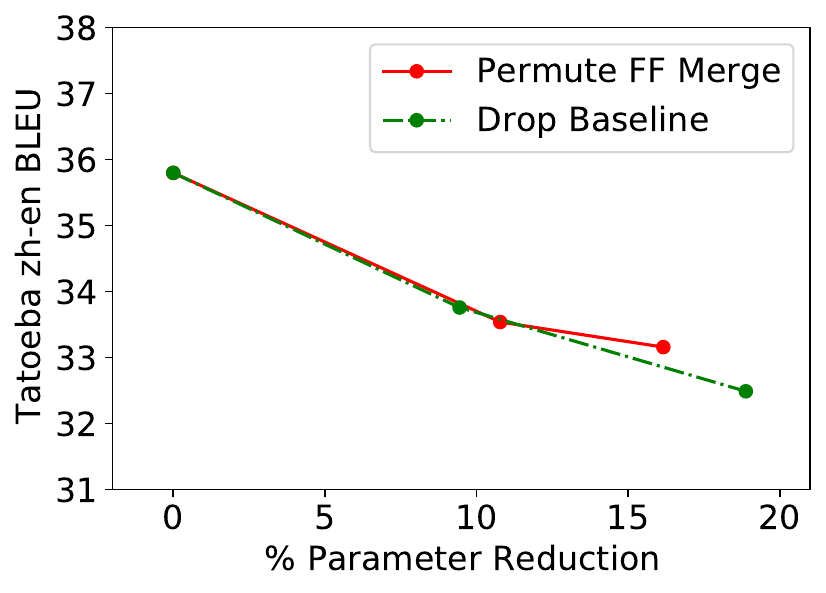} }} \\
    \caption{Results across all three tasks depicting compression versus performance for our method and a strong layer-dropping baseline method. We perform layer dropping for 1/6 and 1/3 of layers dropped, and fine-tune the best pre-tuned set of dropped layers for all sliding windows. Across the parameter reduction range shown, our merging-based compression method outperforms or matches layer-dropping across the three tasks. }
    \label{fig:baselined}
\end{figure*}

In Figure \ref{fig:baselined}, we compare our method at 1/3 and 1/2 FFs removed to our layer-pruning baseline.\footnote{These reduction ratios reflect common ratios found in layer-pruning literature.} We drop layers to attempt to match the reduction ratios of our own methods, constituting 1/6 and 1/3 of layers dropped for all three models. However, since we cannot match exact ratios due to the granularity of the methods, we plot the exact parameter reduction ratios and performance, and compare. As seen in the figure, our method consistently matches or outperforms the layer-dropping method. This comparison confirms that merging is a competitive alternative to strong pruning-based methods for model compression.

\subsection{Choice of merged sublayers}

\begin{figure*}[h!]%
    \centering
    \subfloat[\centering ViT]{{\includegraphics[width=0.3\textwidth]{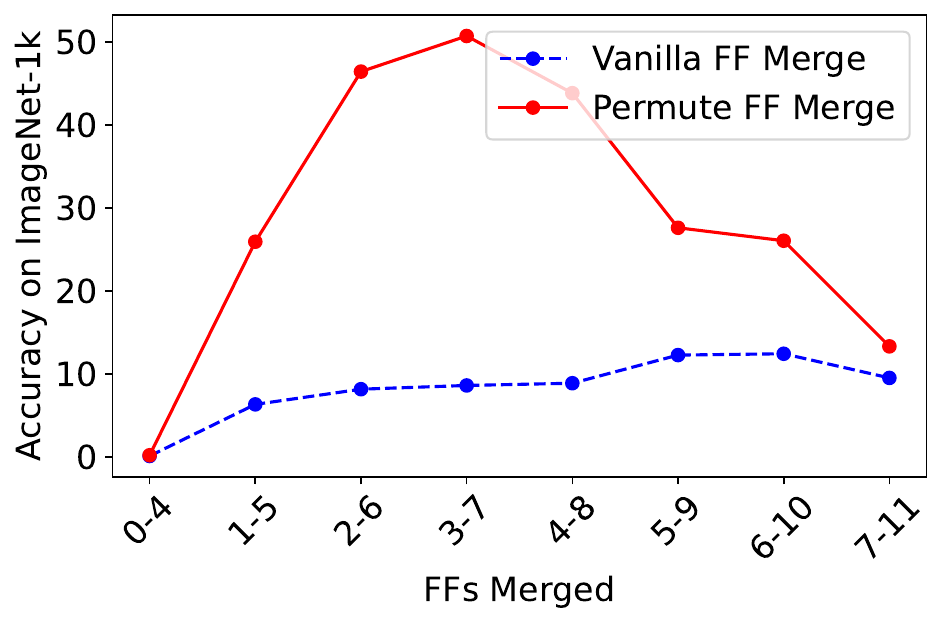} }}
    \subfloat[\centering GPT-2]{{\includegraphics[width=0.3\textwidth]{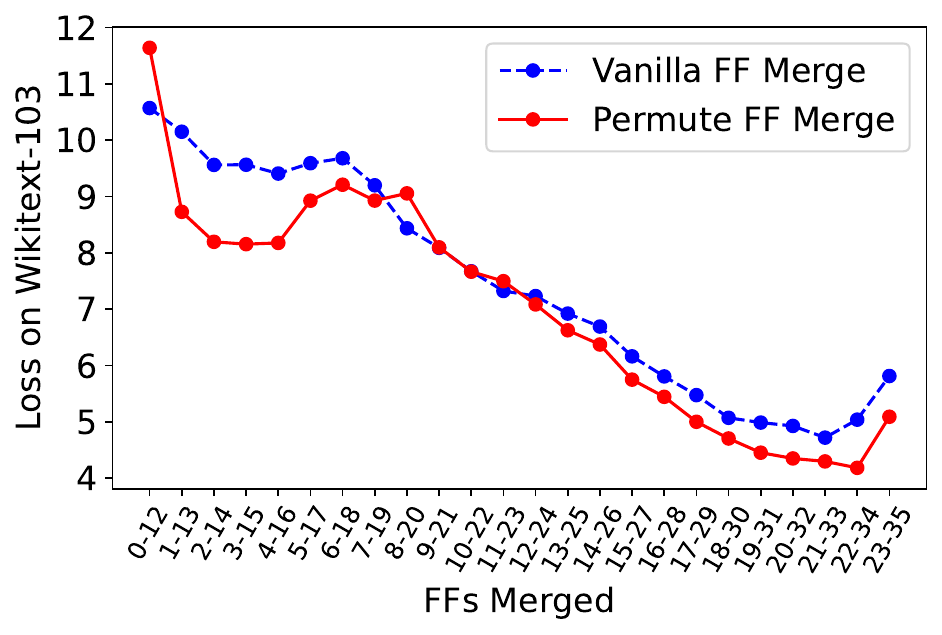} }}%
    \subfloat[\centering OPUS-MT]
    {{\includegraphics[width=0.3\textwidth]{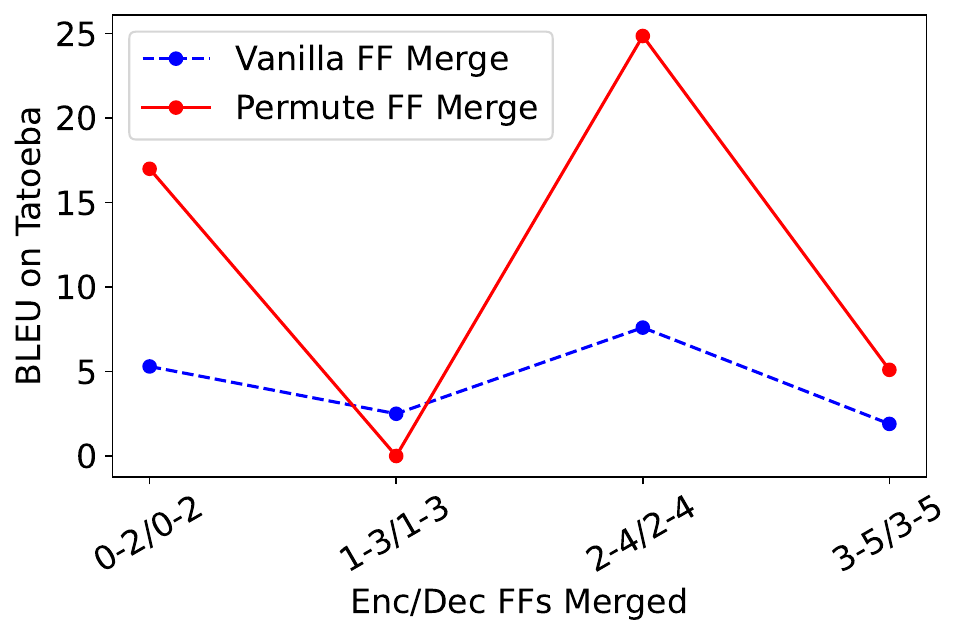} }} \\
    \caption{Performance curves over different ranges of merged feed-forward sublayers representing 1/3 FFs removed. Across all three tasks, there are clear ranges of merged sublayers that retain more performance when merged.\protect\footnotemark~}
    \label{fig:layer_group_curves}
\end{figure*}

In our merging algorithm, we choose which layers to merge by computing performance over sliding windows of $k$ indices. 
For each of our model/task pairs, we plot the pre-tuning performance of the merging algorithm on 1/3 of FF sublayers dropped across all windows, to observe their differences. Results are shown in Figure \ref{fig:layer_group_curves}. 
Before tuning, it appears that the choice of layers seems to be important, resulting in different performance. 
However, these differences reduce once recovery fine-tuning is performed. 
To see this, we randomly select 3 sets of $k$ consecutive layers for each of our tasks, and apply recovery fine-tuning to these compressed models. In Table \ref{table:post-tune_rand}, we observe that all models achieve similar performance after fine-tuning. 

This suggests that the aligning, tying, and tuning portions of our method drive much of the performance improvement. Nevertheless, the choice of layers might be more important in other models; this is potential future work. 

\begin{table}[h!]
\centering
\resizebox{\columnwidth}{!}{
\begin{tabular}{@{}lccc@{}}
\toprule
        & ViT                     & GPT-2            & OPUS-MT         \\
      & Accuracy(\%) $\uparrow$ & PPL $\downarrow$ & BLEU $\uparrow$ \\
\midrule
Best pre-tune & 79.2                    & 17.3             & 33.5            \\
Random 1      & 79.5                    & 18.3             & 33.9            \\
Random 2      & 78.5                    & 17.1             & 33.8            \\
Random 3      & 78.9                    & 17.3             & 33.1            \\
\bottomrule
\end{tabular}
}
\caption{Results comparing our compression method at 1/3 of feed-forward sublayers removed with different sublayer groups. We include three random consecutive selections of sublayers, excluding the original selection.}
\label{table:post-tune_rand}
\end{table}

\subsection{Choice of anchor layer}

We also examine the sensitivity of our method to the choice of anchor layer in our alignment step. 
In section \ref{sec:main_method}, we choose the first FF sublayer in the sequence to serve as the anchor, and compute permutations aligning the remaining sublayers to the anchor. 
Here, we also consider using either the \textit{last} FF or the \textit{middle} FF of the sequence, and report results in our 1/3 FF merge setting in Table \ref{table:anchor}.

\begin{table}[h]
\centering
\resizebox{\columnwidth}{!}{
\begin{tabular}{@{}lccc@{}}
\toprule
     & ViT                     & GPT-2            & OPUS-MT         \\
    & Accuracy(\%) $\uparrow$ & PPL $\downarrow$ & BLEU $\uparrow$ \\
\midrule
Anchor: First & 79.2                    & 17.3             & 33.5            \\
Anchor: Middle & 79.5                    & 17.4             & 33.4            \\
Anchor: Last   & 79.0                    & 17.4             & 33.5             \\
\bottomrule
\end{tabular}
}
\caption{Results comparing our compression method with 1/3 of feed-forward sublayers removed, but with different anchor locations.}
\label{table:anchor}
\end{table}

Given the similar results across settings, our merging approach is robust to the choice of reference or anchor layer, enhancing the reliability of our permutation-based alignment method to find corresponding features for a useful merge.

\begin{table*}[t!]
\centering
\resizebox{0.6\paperwidth}{!}{
\begin{tabular}{@{}llcccc@{}}
\toprule
& &  \multicolumn{2}{c}{Our Method} & \multicolumn{2}{c}{+LLM.int8()} \\
\cmidrule(lr){3-4} \cmidrule(lr){5-6}
Model & Metric &  Compression & Performance & Compression & Performance \\ \midrule
ViT & Accuracy(\%) $\uparrow$ & 78\% &  79.2 & 20\% &  79.2 \\
 GPT-2  & PPL $\downarrow$& 80\% &  17.3 & 22\% &  17.3 \\
 OPUS-MT & BLEU $\uparrow$  &  89\% & 33.5 & 51\% & 33.5 \\
\bottomrule
\end{tabular}
}
\caption{Compression results across three tasks, before and after additional compression via quantization. In this case, compression is measured in terms of total model storage complexity (disk space) instead of parameter count.}
\label{table:quant}
\end{table*}

\footnotetext{We display loss on Wikitext-103 for visibility.  }

 \begin{figure*}[h!]%
    \centering
    \subfloat[\centering ViT]
    {{\includegraphics[width=0.30\textwidth]{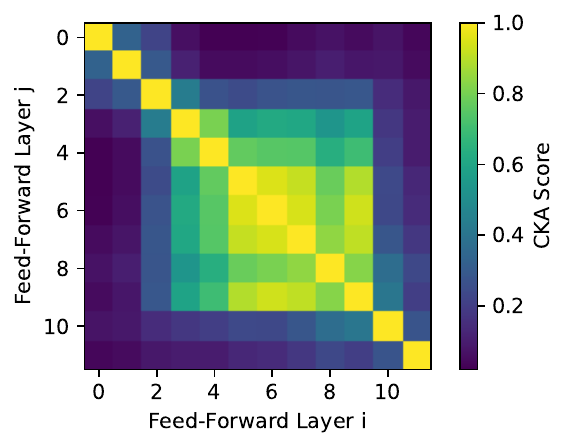} }} 
    \subfloat[\centering GPT2]{{\includegraphics[width=0.30\textwidth]{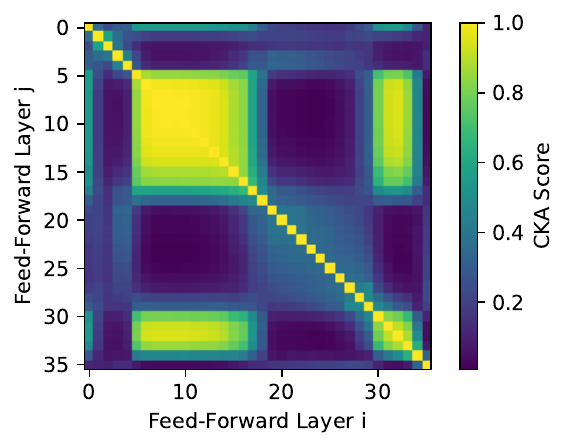} }}%
    \subfloat[\centering OPUS-MT]{{\includegraphics[width=0.30\textwidth]{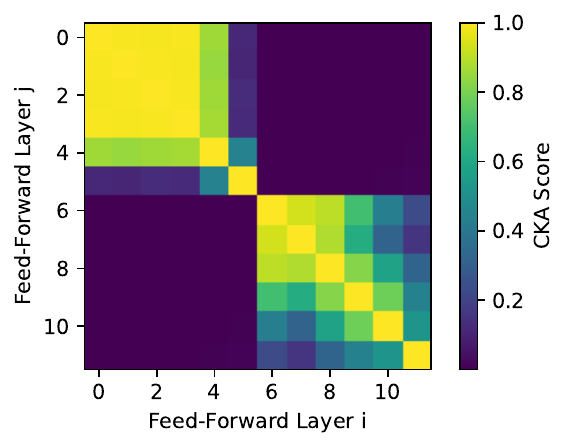} }}\\
    \caption{CKA plots of feed-forward sublayer hidden states across three different models. In all three settings, we see clear regions of high similarity between different FF layers. }%
    \label{fig:cka}%
\end{figure*}

\subsection{Additional compression via quantization}

While our compression method reduces model size via weight tying, quantization also reduces the overall storage needed for a model via reducing parameter precision.
If our method performs orthogonally to state-of-the-art quantization, both methods may be used together for additional memory savings. We experiment with the LLM.int8() quantization method due to its effectiveness and widespread adoption \citep{dettmers2022gpt3}. 

We quantize our models after removing 1/3 of FF  sublayers, and report results in Table \ref{table:quant}. Combining our method with quantization provides even smaller compression ratios, while retaining high performance. 
This coupling helps to realize compression ratios like 20\% when considering total model storage complexity. 

\subsection{QLoRA Extension on OLMo-7B}

In combining our method with QLoRA, we are able to reduce the size of the base model further before quantizing and fine-tuning on SamSum. We report results on the base model, layer pruning, and our method at 1/3 FFs merged in Table \ref{table:qlora}.

\begin{table}[]
\centering
\resizebox{\columnwidth}{!}{
\begin{tabular}{lcccc}
\toprule
  & Layers & ROUGE-1 & ROUGE-2 & ROUGE-Lsum \\ \midrule
\textit{QLoRA} & - & 53.68 & 29.34 & 49.83\\
~~~+Drop  & 10-16 & 50.95 & 26.82&47.28 \\
~~~+Merge & 9-20 & 51.61 & 26.95 & 47.41 \\
\bottomrule 
\end{tabular}
}
\caption{Summarization results reported in ROUGE scores for the SamSum dataset. Our merging method can help reduce the base model further while still allowing for high downstream summarization performance.}
\label{table:qlora}
\end{table}

While our method slightly underperforms the base setting, we slightly improve over the Drop baseline setting and maintain high downstream performance. We note that for both the layer pruning baseline as well as our method, the base model is unchanged after compressing, which demonstrates that our method holds up to this aspect of LoRA \cite{hulora}, and can aid in settings where base models need to be reduced further before tuning.

\subsection{Similarity trends across feed-forward sublayers}

Given the success of simply aligning and merging adjacent feed-forward sublayers for compression, we look further into possible signs of redundancy in their representations, as alluded to in previous work \citep{pires-etal-2023-one, kobayashianalyzing}. 

To this end, we compare outputs between FF sublayers within the same models. Across our three tasks, we use 10k tokens or patches from task validation sets to compute output states from all feed-forward sublayers. 
Then, we use Centered Kernel Alignment (CKA) to compute their similarity \citep{kornblith2019similarity}. 
We plot CKA values for all pairwise interactions between FF sublayers in all three of our model types, shown in Figure \ref{fig:cka}.

We notice that across all three model/task pairs, clear regions of high similarity between FF outputs can be observed, despite FF sublayers being interleaved with multi-headed attention sublayers. We note that similar behavior is not seen in attention sublayers, as seen in Appendix \ref{app:attn_sim}. 
While prior work has shown high similarity between the outputs of adjacent \textit{full} Transformer layers, that result can be explained in part by the residual computations that add prior sublayer outputs to current sublayer outputs \citep{kornblith2019similarity, dalvi-etal-2020-analyzing}. However, here we isolate the FF outputs from the stream of residual computations, before this output is added back in, making the observed similarity more surprising due to the greater independence between FF computations.

\section{Conclusion}

In this work, we propose a novel compression method for Transformer models via merging and tying adjacent sets of FF sublayers. 
Our method explores an alternative to existing compression approaches, and opens possibilities of future methods that use  parameter merging and tying as a post-training compression technique. 
We demonstrate our method's extensibility across diverse tasks, and show that it helps retain high performance even after removing 1/3 of FF sublayers, and outperforms a strong layer pruning baseline. 
Our method can be combined with quantization and QLoRA to achieve even smaller compression ratios, which can help open opportunities for model use across more hardware. 
Finally, we find that several FF sublayers activate very similarly despite being separated by attention sublayers, which may be related to their surprising mergeability.

\section{Limitations}

This merging-based compression method sits between many unstructured pruning methods and structured pruning methods, where the former generally does not result in speed-up or easily realized compression, but the latter can more easily lead to both speed-up and easily realized compression. This work leads to easily realizable compression on disk or GPU, and indirect speed-ups given batch size increases or other optimizations related to parameter on- or off-loading. However, as we do not focus on achieving inference speed-up, this is a limitation of our work.

Additionally, our method is designed and tested on models that use a Transformer-based architecture. While weight-tying and neuron alignment may apply straightforwardly to other architectures, we do not test this, which constitutes another limitation of this work.


\bibliography{acl_latex}

\newpage
\appendix

\section{Combining SwiGLU Feed-Forward Layers }
\label{app:swiglu}

The SwiGLU variation of the Transformer FF layer is used across numerous current language models, including OLMo \cite{groeneveld2024olmo}. The SwiGLU FFN is: 
\begin{equation}
    \text{FF}_{\text{SwiGLU}} =  W^{\text{down}} (\text{Swish}_1(W^{\text{up} }x) \otimes  V^{\text{gate}}x ) 
\end{equation}
where $\otimes$ is the component-wise product. We exclude biases here for simplicty and lack of inclusion in OLMo. For more details, we refer the reader to \citet{shazeer2020glu}. In applying our method to SwiGLU FFs, we note three things: 1) the location of our feature collection is the output of the component-wise product in order to find the best permutation for both $W^{\text{up}}$ and $V^{\text{gate}}$, and 2) the new merged parameters are computed as:
\begin{align}
    W^{\text{up}*} &= \frac{1}{k} \left(W_0^{\text{in}} + \sum_{i=1}^{k-1} P_i W_i^{\text{in}} \right)  \\
    V^{\text{gate}*} &= \frac{1}{k}\left(V_0^{\text{gate}} + \sum_{i=1}^{k-1} P_i V_i^{\text{gate}} \right) \\
    W^{\text{down}*} &= \frac{1}{k} \left(W_0^{\text{down}} + \sum_{i=1}^{k-1} W_i^{\text{down}}P_i^T\right) 
\end{align}

\section{Layer Selection Algorithm}

We summarize our layer selection + fine-tuning algorithm from Section \ref{sub:select_layers} in Algorithm \ref{alg:ff_merge}. 
\label{app:algo_select}
\begin{algorithm}[h]
   \caption{Feed-Forward Sublayer Merge}
   \label{alg:ff_merge}
\begin{algorithmic}
   \STATE {\bfseries Input:} Model parameters $\theta_{\text{in}}$, collected features $\{X_i\}_{i=0}^{N_{\text{layers}} - 1}$, batched fine-tuning data $D_{\text{ft}}$
   \STATE {\bfseries Input constants:} $k$, $N_{\text{layers}}$, \textsc{MaxUpdates}
   \STATE {\bfseries Initialize:}  $\theta_{\text{selected}}$, \textsc{BestScore} $ \gets 0$ 
   \FOR{$i=0$ {\bfseries to} $(N_{\text{layers}} - 1)-k$}
   \STATE $\theta_{\text{merged}} \leftarrow$ \textsc{Compress}$(\theta_{\text{in}}, \{X_i\}_{i=0}^{N_{\text{layers}} - 1}, k )$ 
   \IF {\textsc{Eval}($\theta_{\text{merged}}$) > \textsc{BestScore}}
    \STATE $\theta_{\text{selected}} \leftarrow \theta_{\text{merged}}$
    \ENDIF
   \ENDFOR  
   \FOR{$i=0$ {\bfseries to} \textsc{MaxUpdates}}
   \STATE $\theta_{\text{selected}} \leftarrow$ \textsc{Update}($\theta_{\text{selected}}, D_{\text{ft}}(i)$) 
   \ENDFOR
   \STATE {\bfseries Output:} $\theta_{\text{selected}}$
\end{algorithmic}

\end{algorithm}

\section{Fine-tuning details}
\label{app:ft_details}
\subsection{GPT-2}

\label{app:ft_details_gpt2}

\begin{table}[H]
\centering
\begin{tabular}{@{}lc@{}}
\toprule
Hyperparameter & Value \\ \midrule
Start LR & 5e-5 \\
LR Schedule & \texttt{inv\_sqrt}\\
fp16 & True \\
batch size & 2 \\
n\_steps & 100K \\
\bottomrule
\end{tabular}
\caption{Hyperparameters used for GPT-2 fine-tuning}
\label{table:gpt2_hparam}
\end{table}

\subsection{ViT}

\label{app:ft_details_vit}

\begin{table}[h!]
\centering
\begin{tabular}{@{}lc@{}}
\toprule
Hyperparameter & Value \\ \midrule
Start LR & 5e-5 \\
LR Schedule & \texttt{lin\_decay with min}\\
decay\_steps & 20K \\
Min LR & 1e-6 \\
fp16 & True \\
batch size & 128 \\
n\_steps & 50K \\
\bottomrule
\end{tabular}
\caption{Hyperparameters used for ViT fine-tuning }
\label{table:vit_hparam}
\end{table}

\subsection{Machine Translation}

\label{app:ft_details_mt}

We select our best model using validation BLEU, computed on a 2000 instance subset of the full Tatoeba validation set. 

\begin{table}[h!]
\centering
\begin{tabular}{@{}lc@{}}
\toprule
Hyperparameter & Value \\ \midrule
Start LR & 5e-5 \\
LR Schedule & \texttt{inv\_sqrt}\\
fp16 & True \\
batch size & 64 \\
n\_steps & 100K \\
\bottomrule
\end{tabular}
\caption{Hyperparameters used for OPUS-MT fine-tuning }
\label{table:mt_hparam}
\end{table}

\subsection{OLMo-7B QLoRA}
\label{app:olmo_ft_details}

We select our best model using validation loss. 
\begin{table}[h!]
\centering
\begin{tabular}{@{}lc@{}}
\toprule
Hyperparameter & Value \\ \midrule
Start LR & 5e-4 \\
LR Schedule & \texttt{inv\_sqrt}\\
warmup ratio  & 0.01 \\
fp16 & True \\
batch size & 8 \\
n\_steps & 3000 \\
weight decay & 0.01 \\
LoRA rank & 8 \\
LoRA $\alpha$ & 16 \\
LoRA modules & all linear \\
LoRA dropout & 0.2 \\
\bottomrule
\end{tabular}
\caption{Hyperparameters used for OLMo-7B QLoRA fine-tuning }
\label{table:mt_hparam}
\end{table}

\newpage
\section{Dataset details}
\label{app:data_details}

We report the dataset statistics for our evaluations and training data used in this work in Table \ref{table:data_details}. For fine-tuning data, we note that updates reported in Appendix \ref{app:ft_details} give a representation of data usage rather than the training counts provided here. 
\begin{table}[h!]
\centering
\begin{tabular}{@{}lrrr@{}}
\toprule
Dataset & Train & Validation & Test \\ \midrule
ImageNet-1k & 12,281,167 & 50,000 & -\\
Wikitext-103 & 1,801,350 & 3,760 & 4,358 \\
OPUS/Tatoeba & 41,649,946 & 43,074 & 10,389 \\
SamSum & 14,732 & 818 & 819 \\
\bottomrule
\end{tabular}
\caption{The number of instances used in each fine-tuning and evaluation datasets. Instances are image for ImageNet, lines of text for Wikitext-103, bitext pairs for OPUS/Tatoeba, and dialogue/summary pairs for SamSum.}
\label{table:data_details}
\end{table}

\section{Full Results at varying compression  ratios}
\label{app:full_results}
We report our full results across compression ratios in Table \ref{table:full_results}.

\begin{table*}[t!]
\centering
\begin{tabular}{@{}lllccc@{}}
\toprule
\textbf{Model} & \textbf{Metric} & \textbf{Merged Indices} & \textbf{FFs Removed} & \textbf{Vanilla} & \textbf{Permute}  \\
\midrule 
\multirow{4}{*}{ViT}&\multirow{4}{*}{Accuracy (\%) $\uparrow$}  & -- & 0/12 & 80.3 & 80.3\\
& & 3-7 & 4/12 & 77.8 & 79.2 \\
& & 4-10 & 6/12 & 75.3  & 76.3 \\
& & 0-11 & 11/12&  39.0 & 58.1 \\
\midrule
\multirow{4}{*}{GPT-2}&\multirow{4}{*}{PPL $\downarrow$} & -- & 0/36 & 16.16 & 16.16 \\
& & 22-34 & 12/36 &  17.39 & 17.27 \\
& & 16-34 & 18/36 & 19.01& 18.66\\
& & 0-35 & 35/36 & 23.02 & 21.31\\
\midrule
\multirow{4}{*}{OPUS-MT}&\multirow{4}{*}{BLEU $\uparrow$} & -- & 0/12 & 35.8 & 35.8 \\
& & 2-4/2-4 & 4/12 & 33.3 & 33.5 \\
& & 0-3/0-3 & 6/12 & 32.8 & 33.2 \\
& & 0-5/0-5 & 11/12 & 29.3 & 30.1 \\
\bottomrule
\end{tabular}
\caption{Full numerical results on compression results at 1/3 FF sublayers removed, 1/2 FF sublayers removed, and $(n-1)/n$ FF sublayers removed. Original, uncompressed models are included in the first row of results for each model, indicated by 0 FFs removed and no merged indices.}
\label{table:full_results}
\end{table*}

\section{COMET evaluation of MT experiments}
We report the COMET scores of our main experiments, split by section, in Table \ref{table:comet}.
\begin{table*}[t!]
\centering
\begin{tabular}{@{}lllccc@{}}
\toprule
\textbf{Experiment}  & \textbf{Merged Indices} & \textbf{Anchor} & \textbf{FFs Removed} & \textbf{Vanilla} & \textbf{Permute}  \\
\midrule
\multirow{4}{*}{Main}& -- & First &  0/12 & 86.8 & 86.8 \\
& First &2-4/2-4 & 4/12 & 85.7 & 85.7 \\
& First &0-3/0-3 & 6/12 & 85.2 & 85.4 \\
& First &0-5/0-5 & 11/12 & 83.1 & 83.5 \\
\midrule
\multirow{3}{*}{Layer choice}& First & 0/2-0/2  &  4/12 & - & 85.8 \\
& First &1-3/1-3 & 4/12 & - & 85.8 \\
& First &3-5/3-5& 4/12 & - & 85.6 \\
\midrule 
\multirow{2}{*}{Anchor choice}& Middle & 2-4/2-4  &  4/12 & - & 85.8 \\
& Last &2-4/2-4 & 4/12 & - & 85.7 \\
\midrule
+Quantization & First & 2-4/2-4  &  4/12 & - & 85.8 \\
\bottomrule
\end{tabular}
\caption{Comet scores corresponding to BLEU scores in each table. The first section corresponds to Table \ref{app:full_results}, the second Table \ref{table:post-tune_rand}, the third Table \ref{table:anchor}, and the last Table \ref{table:quant}.}
\label{table:comet}
\end{table*}

\section{Attention Layer Similarity}

\begin{figure*}[h]%
    \centering
    \subfloat[\centering ViT]
    {{\includegraphics[width=0.33\textwidth]{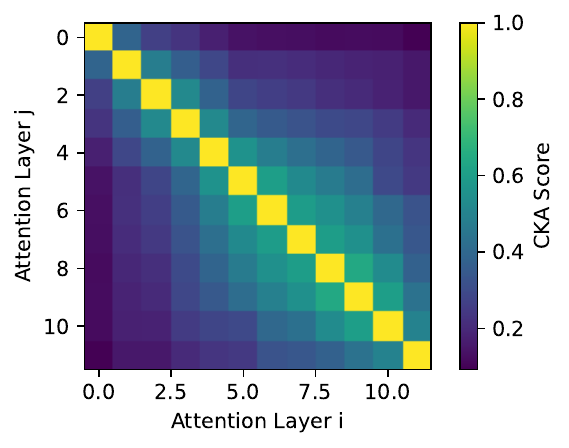} }}
    \subfloat[\centering GPT2]{{\includegraphics[width=0.33\textwidth]{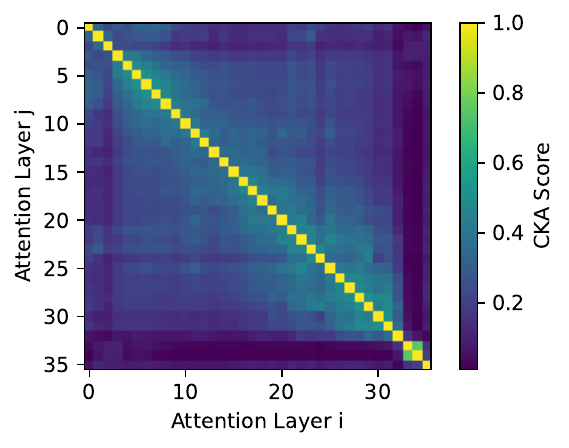} }}%
    \subfloat[\centering OPUS-MT]{{\includegraphics[width=0.33\textwidth]{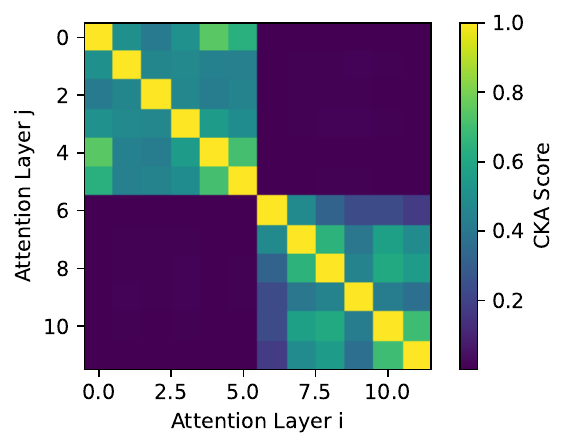} }}
     \\
    \caption{CKA plots of multi-headed self-attention sublayer activations across three different trained models. Attention activations are largely dissimilar from each other across model types.  We do not compare between encoder and decoder attention sublayers in the translation model due the differences in token inputs.}
    \label{fig:attn_cka}%
\end{figure*}

We compute CKA similarity between all attention sublayer pairs, using the same 10k tokens or patches from our CKA results on FF sublayers. 
The features are from the output of the linear layer just after the dot-product attention computation. Results appear in Figure \ref{fig:attn_cka}. 

\label{app:attn_sim}

\end{document}